\documentclass{article}
\usepackage{spconf,amsmath,graphicx}
\usepackage{amsfonts}
\usepackage{algorithmic}
\usepackage{textcomp}
\usepackage{xcolor}
\usepackage{framed,multirow}
\usepackage{multicol,setspace}
\usepackage{multirow}
\usepackage{makecell}
\usepackage{tabularx}
\usepackage{soul}

\title{Multi-head Temporal  Attention-Augmented Bilinear Network\\for Financial time series prediction}

\name{Mostafa Shabani$^{\dagger}$ \quad Dat Thanh Tran$^{\star}$ \quad Martin Magris$^{\dagger}$ \quad Juho Kanniainen$^{\star}$ \quad Alexandros Iosifidis$^{\dagger}$}
\address{$^{\dagger}$ Department of Electrical and Computer Engineering, Aarhus University, Denmark \\$^{\star}$ Unit of Computing Sciences, Tampere University, Finland}

\begin{document}
%
\maketitle
\begin{abstract}
Financial time-series forecasting is one of the most challenging domains in the field of time-series analysis. This is mostly due to the highly non-stationary and noisy nature of financial time-series data. With progressive efforts of the community to design specialized neural networks incorporating prior domain knowledge, many financial analysis and forecasting problems have been successfully tackled. The temporal attention mechanism is a neural layer design that recently gained popularity due to its ability to focus on important temporal events. In this paper, we propose a neural layer based on the ideas of temporal attention and multi-head attention to extend the capability of the underlying neural network in focusing simultaneously on multiple temporal instances. The effectiveness of our approach is validated using large-scale limit-order book market data to forecast the direction of mid-price movements. Our experiments show that the use of multi-head temporal attention modules leads to enhanced prediction performances compared to baseline models.
\end{abstract}
\section{Introduction}
Time-series analysis has been significantly improved by recent machine learning and deep learning approaches. One of the most challenging domains in time-series analysis is that financial time-series classification and prediction. The complex dynamics of financial markets reflect in highly non-stationary and noisy data. This characteristic and the large-scale high-dimensional nature of financial data strongly affect the analysis of financial time-series data. To tackle challenges in financial time-series analysis, many approaches have been proposed based on econometric, machine learning, and deep learning techniques.  

In recent years, the accessibility to large-scale datasets and the improvements in computational capabilities have enabled deep Learning to excel in a variety of domains such as computer vision and natural language processing. 
Popular neural network designs for financial time-series include Recurrent Neural Networks (RNN) \cite{mandic2001recurrent}, of which the Long-Short Term Memory (LSTM) \cite{fischer2018deep} and the Gated Recurrent Unit (GRU) \cite{cho2014learning} are the most widely used recurrent cells. 
Convolutional Neural Network (CNN) \cite{lecun1998gradient}, which was originally designed for visual data, is nowadays also a popular choice for time-series data.

Recently, neural networks that are designed using multilinear operations have also shown competitive performance in time-series analysis tasks compared to recurrent or convolutional networks \cite{Tran2019a}. The Temporal Attention-Augmented Bilinear (\texttt{TABL}) is a neural network layer based on bilinear projection and attention mechanism that adaptively learn to mask out irrelevant time instances \cite{Tran2019a}. 
A new architectural design called Transformer \cite{vaswani2017attention}, which heavily employs multiple attention modules, has emerged as a state-of-the-art model in language understanding tasks \cite{devlin2018bert}, as well as vision understanding tasks \cite{dosovitskiy2020image}.   

In this paper, inspired by the recent success of multi-head attention design, we propose an extension of the \texttt{TABL} network with multi-head attention design. The new design enables a bilinear mapping with the ability to simultaneously learn to focus on different temporal instances in the input time-series. As a result, more discriminative features can be extracted using our neural layer design, which leads to performance improvements compared to the original \texttt{TABL} networks. The remainder of this paper is organized as follows. In Section \ref{sec:related_works}, we provide a literature review on deep learning research for financial time-series forecasting. In Section \ref{sec:Proposed_method}, we describe the proposed multi-head attention design for bilinear mapping. In Section \ref{sec:experiments}, experimental protocols and empirical results are presented. Section \ref{sec:conclusion} concludes our paper.

\section{Related works}
\label{sec:related_works}
The complex dynamics of financial data and the existence of large-scale datasets have fostered the use of deep learning models in financial applications. Among those, analysis tasks derived from high frequency Limit-Order Book (LOB) data have attracted great attention from the community due to its unique capability in tracking market dynamics. 
A comprehensive description on LOBs can be found in \cite{Bouchaud2002}. 

Since our work focuses on analyzing LOB data, here we review related works in LOB research. There have been several works using LOB data. For example, the spatial distribution in LOB has been studied in \cite{sirignano2019deep} via deep neural networks. The LOB data is generally highly non-stationary and requires great attention in terms of pre-processing. Adaptive data normalization schemes have been proposed recently to tackle such challenges \cite{passalis2019deep, tran2020data}.  Designing suitable neural network architectures for time-series derived from LOB has also been the focus of several works, including both manually \cite{Zhang2019, tran2020attention} and automatically generated network architectures \cite{Tran2019}. Beside recurrent networks and \texttt{TABL} networks, neural networks constructed from Bag-of-Feature layers \cite{passalis2019tetci} have also demonstrated a great fit for variable-length sequences.

Among the human expert designs, attention module has shown consistent ability to enhance the baseline models. The main idea of an attention unit is to learn to focus on relevant parts of the input while mask out unimportant parts of it. Attention computation in neural networks was first introduced for machine translation tasks by the work of \cite{bahdanau2014neural}. Incorporation of attention mechanism is also popular among time-series analysis community \cite{makinen2019forecasting, tran2020attention, qin2017dual, Tran2019a, 9308440}. 
Our work relies on a computationally fast and efficient design called Temporal Attention-augmented Bilinear Layer (\texttt{TABL}) network \cite{Tran2019a}, which has been shown to achieve excellent performance in both computational cost and modeling capacity. To have a better understanding of our proposed multi-head attention design in Section \ref{sec:Proposed_method}, the working mechanism of a \texttt{TABL} is described next.

In \texttt{TABL}, the bilinear projection incorporating a temporal attention mechanism produces an output matrix  $\mathbf{Y} \in \mathbb{R}^{D' \times T'}$ given an input matrix $\mathbf{X} \in \mathbb{R}^{D \times T}$. $\mathbf{X}$ is a multivariate time-series in which each column represents the $D$ features at a certain time instance, for a series of length $T$. 
A \texttt{TABL} layer performs five computational steps to transform the input $\mathbf{X}$ to the output $\mathbf{Y}$ as follows:
\begin{equation}
        \bar{\mathbf{X}} = \mathbf{W}_1 \mathbf{X}. \label{eq:TABLa}
\end{equation}
 \begin{equation}
        \mathbf{E} = \bar{\mathbf{X}} \mathbf{W}, \label{eq:TABLb}
\end{equation}
\begin{equation}
        \alpha_{ij} = \frac{ \exp(e_{ij}) }{ \sum_{k=1}^T \exp(e_{ik}) }, \label{eq:TABLc}
\end{equation}
\begin{equation}
        \tilde{\mathbf{X}} = \lambda (\bar{\mathbf{X}} \odot \mathbf{A}) + (1-\lambda) \bar{\mathbf{X}}, \label{eq:TABLd}
\end{equation}
\begin{equation}
        \mathbf{Y} = \phi\left( \tilde{\mathbf{X}} \mathbf{W}_2 + \mathbf{B} \right). \label{eq:TABLe}
\end{equation}

\section{Temporal Multi-head Attention Bilinear Layer}\label{sec:Proposed_method}
Our proposed neural layer is constructed based on the structure of \texttt{TABL}. The main idea of our design is to augment the bilinear mapping with multiple attention computation units (otherwise called attention heads), which are calculated independently (in parallel). By using multiple attention heads, we hypothesize that for certain input series, the salient features can appear in pairs, triplets or larger subsets, which cannot be captured by a single attention head. Thus, by extending the number of attention heads, we might be able to detect more relevant features that lie within the input data. To reach this goal, the intermediate output in the \texttt{TABL} layer after going through the linear transformation in the first dimension (output of Eq. (\ref{eq:TABLa})) is used as the input of multiple soft attention heads, generating multiple attended features $\tilde{\mathbf{X}}$  as in Eq. (\ref{eq:TABLd}) for each attention head. Therefore, if we consider $K$ attention heads, each of which is associated with a weight matrix $\{\mathbf{W}^{(\textrm{k})}\}, k=\{1,...,K\}$. The output of all attention heads must be combined based on a strategy, which can be, e.g., summation or concatenation. In this paper, we investigate concatenation for combining the outputs of attention mechanisms as the output of each of the attention mechanisms are used without any processing and losing information.

\begin{figure*}[t]
\centering
        \includegraphics[width=0.8\textwidth]{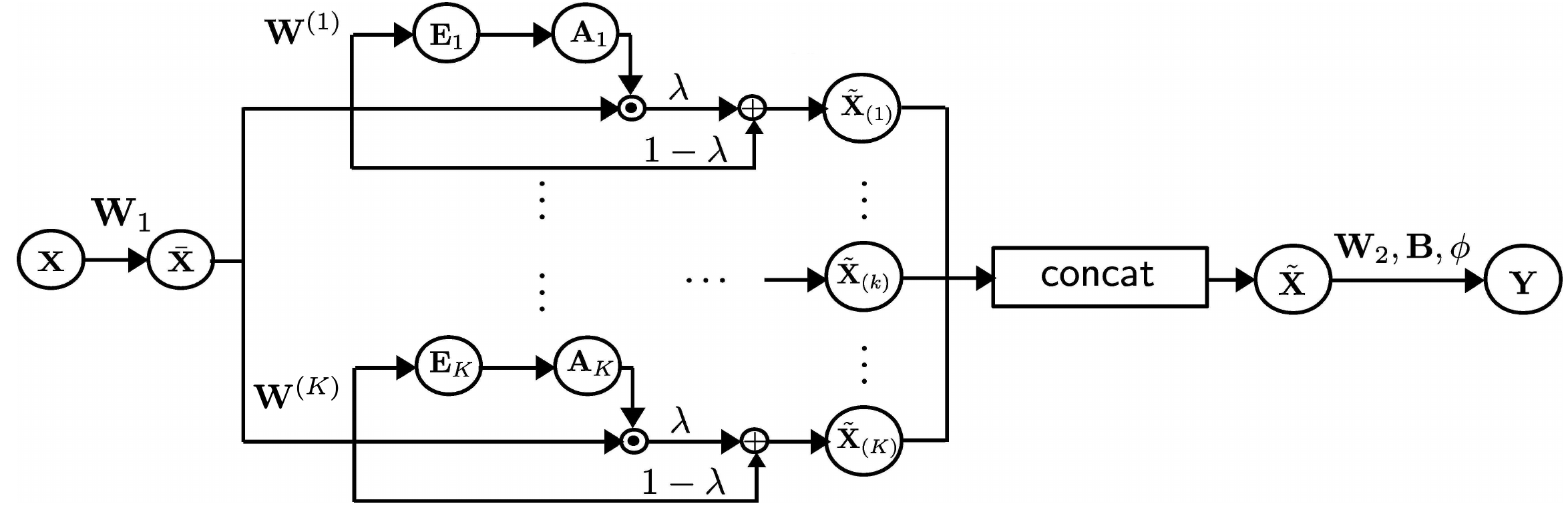}%
        \caption{Schematic illustration of the proposed \texttt{MTABL} layer}
        \label{fig:multi_head_TABL}
\end{figure*}

The computational steps of our Multi-head Temporal Attention Bilinear Layer (\texttt{MTABL}) with $K$ attention heads are as follows:

\begin{itemize}
	\item The first step in \texttt{MTABL} is similar to \texttt{TABL}, which projects each temporal slice (column) of the input matrix to a $D'$-dimensional feature space:
    \begin{equation}
        \bar{\mathbf{X}} = \mathbf{W}_1 \mathbf{X}. \label{eq:mTABLa}
    \end{equation}
        
    \item In the second step, the resulting feature matrix is passed through $K$ parallel attention heads, each of which learns to focus on an important temporal instance:
    \begin{equation}
    \begin{aligned}
        \mathbf{E}_1 = \bar{\mathbf{X}} \mathbf{W}^{(\textrm{1})}, 
        \mathbf{E}_2 = \bar{\mathbf{X}} \mathbf{W}^{(\textrm{2})}, 
       . . . \,,
         \mathbf{E}_K = \bar{\mathbf{X}} \mathbf{W}^{(\textrm{k})}
        \end{aligned}
        \label{eq:MTABLb}
    \end{equation}
    where all $\mathbf{W}^{(\textrm{k})} \in \mathbb{R}^{T\times T}$ is the weight matrix to compute attention in the $k$-th head.
    
    \item The un-normalized attention matrices are then normalized by the softmax function in a row-wise manner, similar to Eq. (\ref{eq:TABLc}), generating the attention masks $\{\mathbf{A}_{(\textrm{k})}\}, k=\{1,...,K\}$
    
    \item The final attended features $\tilde{\mathbf{X}}$ for each attention head are computed by combining the original and masked-out features using the attention mask $\{\mathbf{A}_{(\textrm{k})}\}$ and $\lambda$:
    \begin{equation}
    \begin{aligned}
	    \tilde{\mathbf{X}}_{(\textrm{1})} = \lambda (\bar{\mathbf{X}} \odot \mathbf{A}_{(\textrm{1})}) +  (1-\lambda) \bar{\mathbf{X}}.\\
	    \tilde{\mathbf{X}}_{(\textrm{2})} = \lambda (\bar{\mathbf{X}} \odot \mathbf{A}_{(\textrm{2})}) + (1-\lambda) \bar{\mathbf{X}}.\\
        & \vdots \\
	    \tilde{\mathbf{X}}_{(\textrm{K})} = \lambda (\bar{\mathbf{X}} \odot \mathbf{A}_{(\textrm{K})}) + (1-\lambda) \bar{\mathbf{X}}.\\
    \end{aligned}
    \label{eq:mTABLd}
    \end{equation}
    $\lambda$, which is constrained to have a value between $[0, 1]$, represents the fraction of original information that is relevant and should be allowed to flow through the network when combining with attended features. For this reason, it is more intuitive to use a single value of $\lambda$ for all attention heads. 

    \item All $K$ attended  features $\tilde{\mathbf{X}}_{(\textrm{k})}$ are combined together as a single matrix. To end this, the concatenation is used to combine $\{\tilde{\mathbf{X}}_{(\textrm{k})}\}$. For the concatenation scheme, even though all attended features have the exact same size, it is counterintuitive to concatenate $\{\tilde{\mathbf{X}}_{(\textrm{k})}\}$ on the second dimension, i.e., the temporal dimension, since this means that multiple features of a sequence are concatenated to form a much longer sequence, therefore breaking the temporal coherence of the sequence. Thus, our formulation of method concatenates $\{\tilde{\mathbf{X}}_{(\textrm{k})}\}$ on the feature dimension, then combines all the features of a given temporal instance by linearly projecting them back to $D'$-dimensional space:

	    \begin{equation}\label{eq:con}
		    \tilde{\mathbf{X}} = \tilde{\mathbf{W}}_1 \begin{bmatrix} 
			    \tilde{\mathbf{X}}_{(\textrm{1})}\\
			    \vdots \\

			    \tilde{\mathbf{X}}_{(\textrm{K})}
		    \end{bmatrix} \in \mathbb{R}^{D' \times T}
	    \end{equation}
	where $\tilde{\mathbf{W}}_1 \in \mathbb{R}^{D' \times (D'\cdot K)}$ is a weight matrix that is learned to combine the contenated features.
	
	\item In the final step, similar to \texttt{TABL}, \texttt{MTABL} computes the output sequence $\mathbf{Y}$.
\end{itemize}
Fig. \ref{fig:multi_head_TABL} illustrates the structure of \texttt{MTABL}.

The complexity of \texttt{TABL} is $O(D'DT + D'TT' + 2D'T' + D'T^2 + 3D'T)$~\cite{Tran2019a}. Due to the additional attention heads and the combination of their respective outputs, the \texttt{MTABL} is of greater computational complexity.
In particular, for a \texttt{MTABL} with $K$ attention heads, the number of additional multiplications involved in Eq.~(\ref{eq:MTABLb}) w.r.t  Eq.~(\ref{eq:TABLb}) is $K-1$. Furthermore, an additional complexity term of order $O(D'(D'\cdot K)T)$ is implied by the multiplications in (Eq.~(\ref{eq:con})). The overall complexity of our proposed method is thus $O(D'DT + D'TT' + 2D'T' + KD'T^2 + 3D'T + D'(D'\cdot K)T)$.

\section{EXPERIMENTS}
\label{sec:experiments}
The performance of our model is evaluated on the mid-price movement prediction task using the publicly available FI-2010 dataset \cite{Ntakaris2017}. We used the first $40$ dimensions of the feature vectors, which correspond to the top ten bid and ask prices and volumes of the LOB. For each feature vector, the authors in \cite{Ntakaris2017} derived the labels for future movements of the mid-price in the next $H=\{10, 20, 30, 50, 100\}$ order events, which are referred to as prediction horizons. 

To evaluate the performance of \texttt{MTABL} in comparison to \texttt{TABL}, we used the same experimental protocol of \texttt{TABL} used in \cite{Tran2019a}. We trained all the networks to predict the future movements of mid-price in the next $10$ order events, i.e., the target label corresponding to $H=10$. Three network topologies proposed in \cite{Tran2019a} were used in our experiments as the baseline models The topology $A$ is consist of one \texttt{TABL} layer, the topogy $B$ is consist of one BL layer and one \texttt{TABL} layer and the topology $C$ is consist of two BL layer and one \texttt{TABL} layer.
In these architectures, the last layer is a \texttt{TABL} layer and all other layers are BL layers. We evaluated \texttt{MTABL} networks with varying number of attention heads, from 2 to 5.

\begin{table*}[t!]
	\caption{Performances of multi-head attention models using concatenation (\texttt{MTABL-C}) (Mean $\pm$ STD)}
\label{table:experiments_Concatenation}
\centering
\resizebox{0.8\linewidth}{!}{
\begin{tabular}{|c|c|c|c|c|c|}
\hline
Topology                   & Layer   & Accuracy (\%)     & Precision (\%)    & Recall (\%)       & F1-Score (\%)     \\ \hline
                    & \texttt{TABL}    & 67.21$\pm$0.045 & 53.76$\pm$0.039 & 55.47$\pm$0.015 & 54.25$\pm$0.03  \\ \cline{2-6} 
                    & \texttt{MTABL-C}-2 & 69.78$\pm$0.029 & 56.64$\pm$0.029 & 59.58$\pm$0.026 & 57.81$\pm$0.029 \\ \cline{2-6} 
                    & \texttt{MTABL-C}-3 & 72.45$\pm$0.009 & 59.03$\pm$0.009 & 60.41$\pm$0.001 & 59.66$\pm$0.005 \\ \cline{2-6} 
                    & \texttt{MTABL-C}-4 & 72.30$\pm$0.007 & 59.25$\pm$0.007 & 61.60$\pm$0.005 & 60.28$\pm$0.006 \\ \cline{2-6} 
\multirow{-5}{*}{A} &
  {\color[HTML]{000000} \textbf{\texttt{MTABL-C}-5}} &
  {\color[HTML]{000000} \textbf{72.57$\pm$0.003}} &
  {\color[HTML]{000000} \textbf{59.63$\pm$0.003}} &
  {\color[HTML]{000000} \textbf{62.68$\pm$0.005}} &
  {\color[HTML]{000000} \textbf{60.90$\pm$0.004}} \\ \hline
                    & \texttt{TABL}    & 78.56$\pm$0.002 & 67.55$\pm$0.003 & 71.07$\pm$0.004 & 69.10$\pm$0.002 \\ \cline{2-6} 
                    & \texttt{MTABL-C}-2 & 77.68$\pm$0.004 & 66.44$\pm$0.004 & 70.56$\pm$0.007 & 68.18$\pm$0.004 \\ \cline{2-6} 
                    & \texttt{MTABL-C}-3 & 78.13$\pm$0.007 & 67.04$\pm$0.009 & 71.39$\pm$0.004 & 68.89$\pm$0.007 \\ \cline{2-6} 
                    & \texttt{MTABL-C}-4 & 77.63$\pm$0.005 & 66.48$\pm$0.006 & 70.89$\pm$0.003 & 68.35$\pm$0.005 \\ \cline{2-6} 
\multirow{-5}{*}{B} &
  {\color[HTML]{000000} \textbf{\texttt{MTABL-C}-5}} &
  {\color[HTML]{000000} \textbf{78.22$\pm$0.012}} &
  {\color[HTML]{000000} \textbf{67.4$\pm$0.017}} &
  {\color[HTML]{000000} \textbf{71.52$\pm$0.004}} &
  {\color[HTML]{000000} \textbf{69.16$\pm$0.012}} \\ \hline
                    & \texttt{TABL}    & 83.52$\pm$0.009 & 75.12$\pm$0.013 & 77.02$\pm$0.006 & 76.01$\pm$0.009 \\ \cline{2-6} 
                    & \texttt{MTABL-C}-2 & 83.69$\pm$0.005 & 75.21$\pm$0.008 & 77.74$\pm$0.004 & 76.39$\pm$0.006 \\ \cline{2-6} 
                    & \texttt{MTABL-C}-3 & 81.64$\pm$0.014 & 72.16$\pm$0.022 & 75.17$\pm$0.015 & 73.54$\pm$0.019 \\ \cline{2-6} 
 &
  {\color[HTML]{000000} \textbf{\texttt{MTABL-C}-4}} &
  {\color[HTML]{000000} \textbf{83.71$\pm$0.01}} &
  {\color[HTML]{000000} \textbf{75.37$\pm$0.015}} &
  {\color[HTML]{000000} \textbf{77.63$\pm$0.006}} &
  {\color[HTML]{000000} \textbf{76.42$\pm$0.011}} \\ \cline{2-6} 
\multirow{-5}{*}{C} & \texttt{MTABL-C}-5 & 82.63$\pm$0.004 & 73.66$\pm$0.005 & 76.93$\pm$0.008 & 75.16$\pm$0.006 \\ \hline
\end{tabular}}
\end{table*}

Table \ref{table:experiments_Concatenation} reports the corresponding experiment results for network topologies with concatenation as the attention aggregation strategy to combine attention mechanisms' outputs. Due to the stochastic nature of the optimizer, we report the mean and standard deviation between four independent runs. The following metrics were used to measure the performance of each model: accuracy, precision, recall and F1-Score. Since the FI-2010 dataset has a skewed distribution of labels with the majority of samples having the stationary label, the average F1 score, which reflects the trade-off between precision and recall, is used as the main metric to compare between models. The column ``Layer" indicates which type of output layer was used in the network architecture. The number of attention heads used in each \texttt{MTABL} layer is indicated by the last number in the notation, that is (\texttt{MTABL}-3) denotes a \texttt{MTABL} layer using three attention heads. The whole row corresponds to the model with the best performance for each network topology based on F1-Score is highlighted in bold-face. The results shows the improved performances of the multi-head attention configuration in for all network topologies. This shows that using multiple attentions can help the output layer to detect and focus on crucial elements of data more accurately and improve the prediction performance.

The interpretation of the results from Table \ref{table:experiments_Concatenation} that correspond to network topologies that use concatenation to combine the outputs of all attention heads in each layer is straightforward.
\texttt{MTABL} show the major improvements over the original \texttt{TABL} with five attention heads. On first instance, this can be interpreted as a considerable ($K=5$) amount of relevant attention that is neglected in \texttt{TABL} and that \texttt{MTABL}’s increased number of attention layers captures. When additional BL layers are considered in topologies B and C the best performances are achieved under $K=4$ and $K=5$ respectively but the improvement is not significant.
This can indicate that the five attention heads in topology $A$ the attention-relevant information are highly useful to improve the prediction performance. On the other hand, in topologies $B$ and $C$ when additional BL layers are introduced, the best \texttt{MTABL} performance is still observed with higher number of layers(4 and 5) but the improvement is not as much as topology $A$. This shows that there is little temporal attention discarded in \texttt{TABL} that \texttt{MTABL} captures.

\section{CONCLUSION}
\label{sec:conclusion}
In this paper, a new neural layer based on the structure of \texttt{TABL} and the idea of multi-head attention is proposed for financial time-series analysis. We proposed a formulation of the \texttt{TABL} layer that utilizes multiple attention units to focus on different temporal importances. Our \texttt{MTABL} design stands out as a suitable neural layer for addressing numerous forecasting problems over a wide class of time-series characterized by a complex and time-varying dynamics. 

Extensive experiments in forecasting direction of mid-prices movements using limit-order book data show that the proposed \texttt{MTABL} design is indeed capable of unveiling additional layers of relevant predictive significance lodged in the data. The improved \texttt{MTABL} performance is generally achieved for different combination schemes of the attention heads' outputs, for different number of attention heads, and under different network topologies.

\section*{Acknowledgment}
The research received funding from the Independent Research Fund Denmark project DISPA (Project Number: 9041-00004).
\bibliographystyle{IEEEbib}
\bibliography{main}

\begin{thebibliography}{10}

\bibitem{mandic2001recurrent}
Danilo Mandic and Jonathon Chambers,
\newblock {\em Recurrent neural networks for prediction: learning algorithms,
  architectures and stability},
\newblock Wiley, 2001.

\bibitem{fischer2018deep}
Thomas Fischer and Christopher Krauss,
\newblock ``Deep learning with long short-term memory networks for financial
  market predictions,''
\newblock {\em European Journal of Operational Research}, vol. 270, no. 2, pp.
  654--669, 2018.

\bibitem{cho2014learning}
Kyunghyun Cho, Bart Van~Merri{\"e}nboer, Caglar Gulcehre, Dzmitry Bahdanau,
  Fethi Bougares, Holger Schwenk, and Yoshua Bengio,
\newblock ``Learning phrase representations using rnn encoder-decoder for
  statistical machine translation,''
\newblock {\em arXiv preprint arXiv:1406.1078}, 2014.

\bibitem{lecun1998gradient}
Yann LeCun, L{\'e}on Bottou, Yoshua Bengio, and Patrick Haffner,
\newblock ``Gradient-based learning applied to document recognition,''
\newblock {\em Proceedings of the IEEE}, vol. 86, no. 11, pp. 2278--2324, 1998.

\bibitem{Tran2019a}
Dat~Thanh Tran, Alexandros Iosifidis, Juho Kanniainen, and Moncef Gabbouj,
\newblock ``Temporal attention-augmented bilinear network for financial
  time-series data analysis,''
\newblock {\em IEEE Transactions on Neural Networks and Learning Systems}, vol.
  30, pp. 1407--1418, 2017.

\bibitem{vaswani2017attention}
Ashish Vaswani, Noam Shazeer, Niki Parmar, Jakob Uszkoreit, Llion Jones,
  Aidan~N Gomez, Lukasz Kaiser, and Illia Polosukhin,
\newblock ``Attention is all you need,''
\newblock {\em arXiv preprint arXiv:1706.03762}, 2017.

\bibitem{devlin2018bert}
Jacob Devlin, Ming-Wei Chang, Kenton Lee, and Kristina Toutanova,
\newblock ``Bert: Pre-training of deep bidirectional transformers for language
  understanding,''
\newblock {\em arXiv preprint arXiv:1810.04805}, 2018.

\bibitem{dosovitskiy2020image}
Alexey Dosovitskiy, Lucas Beyer, Alexander Kolesnikov, Dirk Weissenborn,
  Xiaohua Zhai, Thomas Unterthiner, Mostafa Dehghani, Matthias Minderer, Georg
  Heigold, Sylvain Gelly, et~al.,
\newblock ``An image is worth 16x16 words: Transformers for image recognition
  at scale,''
\newblock {\em arXiv preprint arXiv:2010.11929}, 2020.

\bibitem{Bouchaud2002}
Jean-Philippe Bouchaud, Marc M{\'e}zard, and Marc Potters,
\newblock ``Statistical properties of stock order books: empirical results and
  models,''
\newblock {\em Quantitative finance}, vol. 2, no. 4, pp. 251--256, 2002.

\bibitem{sirignano2019deep}
Justin~A Sirignano,
\newblock ``Deep learning for limit order books,''
\newblock {\em Quantitative Finance}, vol. 19, no. 4, pp. 549--570, 2019.

\bibitem{passalis2019deep}
N.~{Passalis}, A.~{Tefas}, J.~{Kanniainen}, M.~{Gabbouj}, and A.~{Iosifidis},
\newblock ``Deep adaptive input normalization for time series forecasting,''
\newblock {\em IEEE Transactions on Neural Networks and Learning Systems}, vol.
  31, no. 9, pp. 3760--3765, 2020.

\bibitem{tran2020data}
Dat~Thanh Tran, Juho Kanniainen, Moncef Gabbouj, and Alexandros Iosifidis,
\newblock ``Data normalization for bilinear structures in high-frequency
  financial time-series,''
\newblock in {\em International Conference on Pattern Recognition (ICPR)},
  2020.

\bibitem{Zhang2019}
Zihao Zhang, Stefan Zohren, and Stephen Roberts,
\newblock ``Deeplob: Deep convolutional neural networks for limit order
  books,''
\newblock {\em IEEE Transactions on Signal Processing}, vol. 67, pp.
  3001--3012, 2019.

\bibitem{tran2020attention}
Dat~Thanh Tran, Nikolaos Passalis, Anastasios Tefas, Moncef Gabbouj, and
  Alexandros Iosifidis,
\newblock ``Attention-based neural bag-of-features learning for sequence
  data,''
\newblock {\em arXiv preprint arXiv:2005.12250}, 2020.

\bibitem{Tran2019}
Dat~Thanh Tran, Juho Kanniainen, Moncef Gabbouj, and Alexandros Iosifidis,
\newblock ``Data-driven neural architecture learning for financial time-series
  forecasting,''
\newblock {\em ArXiv}, vol. abs/1903.06751, 2019.

\bibitem{passalis2019tetci}
N.~{Passalis}, A.~{Tefas}, J.~{Kanniainen}, M.~{Gabbouj}, and A.~{Iosifidis},
\newblock ``Temporal bag-of-features learning for predicting mid price
  movements using high frequency limit order book data,''
\newblock {\em IEEE Transactions on Emerging Topics in Computational
  Intelligence}, pp. 1--12, 2018.

\bibitem{bahdanau2014neural}
Dzmitry Bahdanau, Kyunghyun Cho, and Yoshua Bengio,
\newblock ``Neural machine translation by jointly learning to align and
  translate,''
\newblock {\em arXiv preprint arXiv:1409.0473}, 2014.

\bibitem{makinen2019forecasting}
Ymir M{\"a}kinen, Juho Kanniainen, Moncef Gabbouj, and Alexandros Iosifidis,
\newblock ``Forecasting jump arrivals in stock prices: new attention-based
  network architecture using limit order book data,''
\newblock {\em Quantitative Finance}, vol. 19, no. 12, pp. 2033--2050, 2019.

\bibitem{qin2017dual}
Yao Qin, Dongjin Song, Haifeng Chen, Wei Cheng, Guofei Jiang, and Garrison
  Cottrell,
\newblock ``A dual-stage attention-based recurrent neural network for time
  series prediction,''
\newblock {\em arXiv preprint arXiv:1704.02971}, 2017.

\bibitem{9308440}
M.~{Shabani} and A.~{Iosifidis},
\newblock ``Low-rank temporal attention-augmented bilinear network for
  financial time-series forecasting,''
\newblock in {\em 2020 IEEE Symposium Series on Computational Intelligence
  (SSCI)}, 2020, pp. 2156--2161.

\bibitem{Ntakaris2017}
Adamantios Ntakaris, Martin Magris, Juho Kanniainen, Moncef Gabbouj, and
  Alexandros Iosifidis,
\newblock ``Benchmark dataset for mid-price forecasting of limit order book
  data with machine learning methods,''
\newblock {\em Journal of Forecasting}, vol. 37, pp. 852--866, 2018.

\end{thebibliography}

\end{document}